\documentclass[runningheads]{llncs}
\usepackage[T1]{fontenc}
\usepackage{amsmath}
\usepackage{amssymb}
\usepackage{graphicx}

\usepackage{multirow}
\usepackage{makecell}
\usepackage{bm}
\usepackage{dsfont}

\usepackage[algo2e,ruled,lined]{algorithm2e}

\usepackage{wrapfig}

\usepackage{booktabs}
\usepackage{longtable}

\newcommand{\etal}{\textit{et al}.}
\newcommand{\ie}{\textit{i}.\textit{e}.}
\newcommand{\eg}{\textit{e}.\textit{g}.}

\usepackage{pifont}
\newcommand{\cmark}{\ding{51}}%
\newcommand{\xmark}{\ding{55}}%

\begin{document}
\title{Generalized Category Discovery with Clustering Assignment Consistency}
%
%
\author{Xiangli~Yang\inst{1}\orcidID{0000-0003-0546-5440} 
	\and
Xinglin~Pan\inst{2}
\and
Irwin~King\inst{3}\orcidID{0000-0001-8106-6447}
\and
Zenglin Xu\inst{4}\orcidID{0000-0001-5550-6461}
}
\authorrunning{X. Yang et al.}
%
\institute{School of Computer Science and Engineering, University of Electronic Science and Technology of China, Chengdu, China\\
\email{xlyang@std.uestc.edu.cn}
\and
Department of Computer Science, Hong Kong Baptist University, Hong Kong, China\\
\email{csxlpan@comp.hkbu.edu.hk}\\
\and
Department of Computer Science
and Engineering, The Chinese University of Hong Kong, Hong Kong, China\\
\email{king@cse.cuhk.edu.hk}
\and
School of Computer Science and Technology, Harbin Institute of Technology, Shenzhen, China\\
\email{xuzenglin@hit.edu.cn}}
\maketitle              
\begin{abstract}
Generalized category discovery (GCD) is a recently proposed open-world task. Given a set of images consisting of labeled and unlabeled instances, the goal of GCD is to automatically cluster the unlabeled samples using information transferred from the labeled dataset. The unlabeled dataset comprises both known and novel classes. The main challenge is that unlabeled novel class samples and unlabeled known class samples are mixed together in the unlabeled dataset. To address the GCD without knowing the class number of unlabeled dataset, we propose a co-training-based framework that encourages clustering consistency. Specifically, we first introduce weak and strong augmentation transformations to generate two sufficiently different views for the same sample. Then,  based on the co-training assumption, we propose a consistency representation learning strategy, which encourages consistency between feature-prototype similarity and clustering assignment. Finally, we use the discriminative embeddings learned from the semi-supervised representation learning process to construct an original sparse network and use a community detection method to obtain the clustering results and the number of categories simultaneously.
Extensive experiments show that our method achieves state-of-the-art performance on three generic benchmarks and three fine-grained visual recognition datasets. Especially in the ImageNet-100 data set, our method significantly exceeds the best baseline by 15.5\% and 7.0\% on the \texttt{Novel} and \texttt{All} classes, respectively.

\end{abstract}
%
%
%
\section{Introduction}

\begin{figure}[ht]
	\centering
	\includegraphics[width=1.0\linewidth]{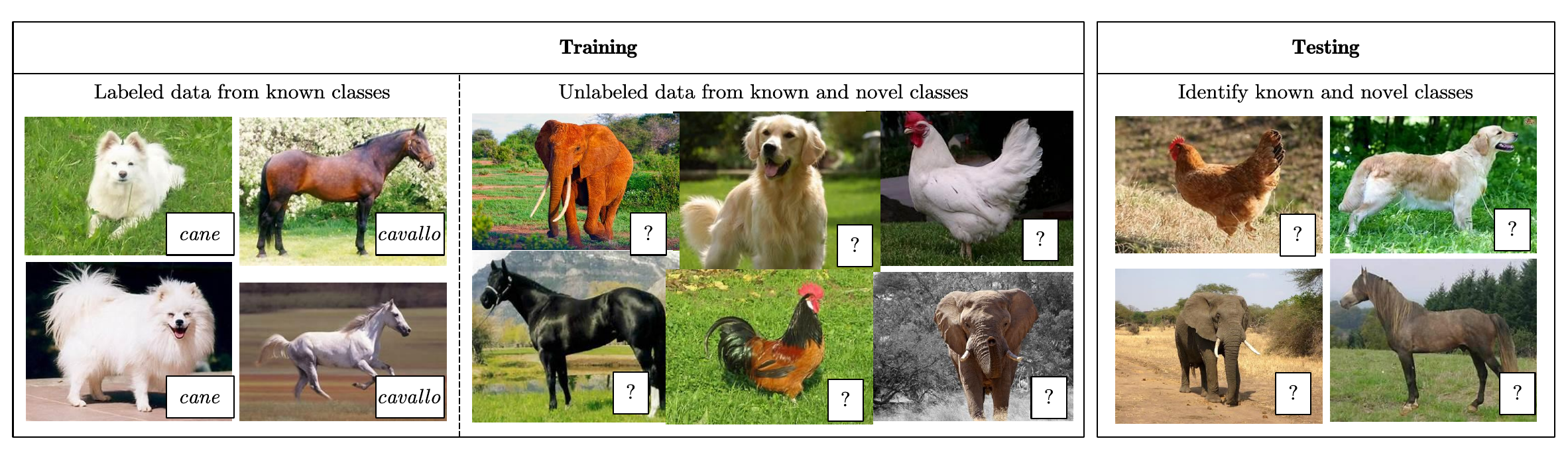}
	\caption{An illustration of the generalized category discovery task. Given a set of labeled images of some known categories, and a set of unlabeled images that contain both known and novel classes,  the objective is to automatically identity known and novel classes in the test dataset.}
	\label{fig:problemsetting}
\end{figure}
\vspace{-8pt}
Deep learning has proven to be remarkably successful in image classification tasks that heavily rely on vast amounts of high-quality labeled data \cite{DBLP:journals/cacm/KrizhevskySH17,DBLP:conf/cvpr/HeZRS16}. However, acquiring annotations for all data is often unfeasible, which requires learning from unlabeled data. Semi-supervised learning \cite{DBLP:conf/nips/BerthelotCGPOR19,DBLP:conf/nips/ZhangWHWWOS21,DBLP:journals/corr/abs-2205-07246} and few-shot learning \cite{DBLP:conf/icml/FinnAL17,DBLP:conf/iclr/DhillonCRS20} are two examples of such learning paradigms. Most of these learning algorithms operate in closed set settings, where known categories are predetermined.

In recent years, \emph{Novel Category Discovery (NCD)} has received significant attention for the discovery of novel classes by transferring knowledge learned from known classes. The implementation of NCD is based on the assumption that all unlabeled samples pertain to novel categories whose number is predetermined, which is excessively idealistic and impractical in the wild. \cite{DBLP:conf/cvpr/VazeHVZ22} relaxes these assumptions by recognizing instances of both known and novel classes in the unlabeled dataset. This new setting is named \emph{Generalized Category Discovery} (GCD). 
To illustrate the setting of GCD, we consider the scenario in Figure~\ref{fig:problemsetting}, where the training set consists of labeled instances of known classes (\eg, cane and cavallo) and unlabeled instances comprising instances of known classes (\ie, cane and cavallo) and novel classes (\ie, gallina and elephant). An ideal model should not only be able to classify known classes (\ie, cane, and cavallo), but also be able to discover novel classes (\ie, gallina, and elephant).

Generalized Category Discovery (GCD) is an open-world problem that holds promise for numerous real-world applications \cite{DBLP:conf/cvpr/VazeHVZ22,DBLP:conf/bmvc/FeiZYZ22,DBLP:journals/corr/abs-2211-15115,DBLP:conf/iclr/CaoBL22}. However, it also poses significant obstacles, namely: a) the lack of supervised prior knowledge for novel classes and b) the absence of clear distinctions between samples belonging to known and novel categories. To address the GCD problem, \cite{DBLP:conf/cvpr/VazeHVZ22} incorporates contrastive loss to learn discriminative representations of the unlabeled dataset. Furthermore, semi-supervised k-means \cite{DBLP:conf/soda/ArthurV07} is applied to implement clustering. However, we find that using all unlabeled instances as negative samples in contrastive learning will treat unlabeled instances of the same category as negative pairs, which will cause category collision problem \cite{DBLP:conf/iccv/Zheng0Y0Z0021}. 
Moreover, in real-world problems, prior information regarding the number of clustering assignments is rarely available, which makes it inappropriate to use semi-supervised k-means.

To overcome these limitations, we present a co-training consistency strategy for clustering assignments to uncover latent representations among unlabeled dataset. For the final clustering goal, we leverage \textit{community detection} techniques to assign labels to unlabeled instances and automatically determine the optimal number of clustering categories based on the learned representations. Our approach
In particular, our approach consists of two stages: semi-supervised representation learning and community detection. In semi-supervised representation learning, we use supervised contrastive loss to derive the labeled information fully. Additionally, since the contrastive learning approach aligns with the co-training assumption, we introduce weak and strong augmentations of the same sample to extract two distinct views. We subsequently deploy the co-training framework to enforce the consistency of feature-prototype similarity and clustering assignment between the two views.  In community detection, we construct the primary graph utilizing feature embeddings learned in semi-supervised representation learning, then apply a community detection method to obtain the outcomes.

Our contributions to this work are as follows,
\vspace{-8pt}
\begin{itemize}
	\item We employ weak and strong data augmentations to generate two sufficiently different views.
	\item We introduce a co-training-based framework to effectively learn embeddings of unlabeled data.
	\item We propose a community detection approach to clustering assignments that accurately detects novel categories of unlabeled instances while simultaneously determining the number of clusters.
	\item Through experimentation on six benchmarks, our approach demonstrates state-of-the-art performance and confirms the efficacy of our method.
\end{itemize}
\vspace{-18pt}

\section{Related Work}
\vspace{-8pt}
\textbf{Novel Category Discovery.} 
The objective of Novel Category Discovery (NCD) is to cluster unlabeled instances using transferable knowledge derived from the labeled dataset of known categories, with all unlabeled instances regarded as novel categories \cite{DBLP:journals/corr/abs-2302-12028}.
KCL \cite{DBLP:conf/iclr/HsuLK18} addresses the problem of transfer learning across domains, especially cross-task transfer learning, which is similar with NCD. Similar to KCL, MCL \cite{DBLP:conf/iclr/HsuLSOK19} constructs pairwise pseudo-labels on unlabeled dataset using a similarity prediction network trained on the labeled dataset, and determines whether two instances belong to the same category by using the inner product. DTC \cite{DBLP:conf/iccv/HanVZ19} utilizes a cross-entropy classifier to initialize representations on the  labeled dataset. Then it maintains a list of class prototypes that represent cluster centers, and assigns instances to the closest prototype. In this way, a good clustering representation is learned.
RankStats or AutoNovel \cite{DBLP:conf/iclr/HanREVZ20,DBLP:journals/pami/HanREVZ22} applies RotNet \cite{DBLP:conf/iclr/GidarisSK18} to perform an initialization of the encoder. Afterward, it generates pseudo-labels through dual-ranking statistics and enforces consistency between the predictions of the two branch networks. NCL \cite{DBLP:conf/cvpr/ZhongFRL0S21} retrieves and aggregates pseudo positive pairs  by contrastive learning, and produces hard negative pairs by mixing labeled and unlabeled instances in the feature space. Unlike the methods that separately design loss functions for labeled and unlabeled samples, UNO \cite{DBLP:conf/iccv/FiniSLZN021} introduces a unified objective function to discover new classes that effectively supports the collaboration between supervised and unsupervised learning. MEDI \cite{DBLP:conf/iclr/ChiLYLL00ZS22} proposes a new approach using MAML \cite{DBLP:conf/icml/FinnAL17} to solve the NCD problem and provided a solid theoretical analysis of the underlying assumptions in the NCD field. ComEx \cite{DBLP:conf/cvpr/YangZYWD22} modifies UNO \cite{DBLP:conf/iccv/FiniSLZN021} by designing two groups of experts to learn the entire dataset in a complementary way, thus mitigating the limitations of previous NCD methods. Recently, GCD \cite{DBLP:conf/cvpr/VazeHVZ22} extends the NCD task to include the recognition of known classes in the unlabeled dataset. It leverages a well-trained visual transformer model \cite{DBLP:conf/iclr/DosovitskiyB0WZ21} to improve visual representation.
ORCA \cite{DBLP:conf/iclr/CaoBL22} enables the simultaneous clustering of various novel classes in the unlabeled set, assuming a pre-defined number of novel classes.

\textbf{Self-supervised learning.} 
Self-supervised learning \cite{DBLP:journals/entropy/Albelwi22} can simultaneously utilize supervised and unsupervised learning during the fine-tuning stage of training without the need for manual annotations \cite{DBLP:conf/cvpr/He0WXG20}. Furthermore, self-supervised learning can be divided into two types: auxiliary pretext learning and contrastive learning. When manual annotations are not available, auxiliary pretext tasks are used as supervised information to learn representations \cite{DBLP:journals/natmi/HolmbergKMSHKAS20}. Some commonly used pretext tasks include exemplar-based methods \cite{DBLP:journals/pami/DosovitskiyFSRB16}, rotation \cite{DBLP:conf/iclr/GidarisSK18}, predicting missing pixels \cite{DBLP:conf/cvpr/PathakKDDE16}, grayscale images \cite{DBLP:conf/eccv/ZhangIE16}, and patch context and jigsaw puzzles \cite{DBLP:conf/iccv/DoerschGE15,DBLP:conf/eccv/NorooziF16}. However, designing a suitable auxiliary pretext task requires domain-specific knowledge in order to better serve downstream tasks.
Contrastive learning minimizes the latent embedding distance between positive pairs and maximizes the distance between negative pairs \cite{DBLP:conf/aaai/YangACX22}.
SimCLR \cite{DBLP:conf/icml/ChenK0H20} learned useful representations based on contrastive loss by maximizing the similarity between the original sample and augmented views of it. MoCo \cite{DBLP:conf/cvpr/He0WXG20} utilized a momentum contrast to calculate the similarity. BYOL \cite{DBLP:conf/nips/GrillSATRBDPGAP20} proposed a new  framework for learning feature representations without the help of contrasting negative pairs. This was achieved by using a Siamese architecture, where the query branch is equipped with a predictor architecture in addition to the encoder and projector. Following BYOL \cite{DBLP:conf/nips/GrillSATRBDPGAP20}, In addition to removing the impulse key encoder, Simsiam \cite{DBLP:conf/cvpr/ChenH21} also used the stop-gradient strategy in the approach while adopting it to alleviate issues related to collapsing. Roh \etal ~\cite{DBLP:conf/cvpr/RohSKK21} utilized consistency losses for the intersection region of the ROIs from two views to improve the encoder representation in subsequent detection tasks. Moreover, the KL loss is a popular approach for consistency learning, as demonstrated in prior works such as \cite{DBLP:conf/iclr/0005W0u21} and \cite{DBLP:conf/iclr/MitrovicMWBB21}, with the incorporation of regularization to ensure that embeddings derived from different data augmentation techniques are consistent.
\vspace{-15pt}

\section{Method}
\vspace{-8pt}
\subsection{Preliminaries}
\vspace{-8pt}
GCD aims to automatically classify unlabeled instances containing both known and unknown categories \cite{DBLP:conf/cvpr/VazeHVZ22}. This is a more realistic open-world setting than the common closed-set classification, which assumes that labeled and unlabeled data belong to the same categories. Our settings follows \cite{DBLP:conf/cvpr/VazeHVZ22}.
Formally, let the train dataset be $\mathcal{D}=\mathcal{D}_{L} \cup \mathcal{D}_{U}$, where $\mathcal{D}_{L}=\{(\bm{x_i},y_i)|y_i\in \mathcal{Y}_{known}\}$ and $\mathcal{D}_{U}=\{\bm{x}_i|y_i\in \mathcal{Y}_U\}$ and $\mathcal{Y}_U=\{\mathcal{Y}_{known}, \mathcal{Y}_{novel}\}$. Here, $\mathcal{Y}_U, \mathcal{Y}_{known}$ and $\mathcal{Y}_{novel}$ denote the label set of \texttt{All}, \texttt{Known}, and \texttt{Novel} classes, respectively. This formalization enables us to easily distinguish between the NCD setting and GCD setting. In NCD, it is assumed that $\mathcal{Y}_{known}\cap \mathcal{Y}_U=\emptyset$. 

\vspace{-15pt}
\begin{figure}[ht]
	\centering
	\includegraphics[width=1.0\linewidth]{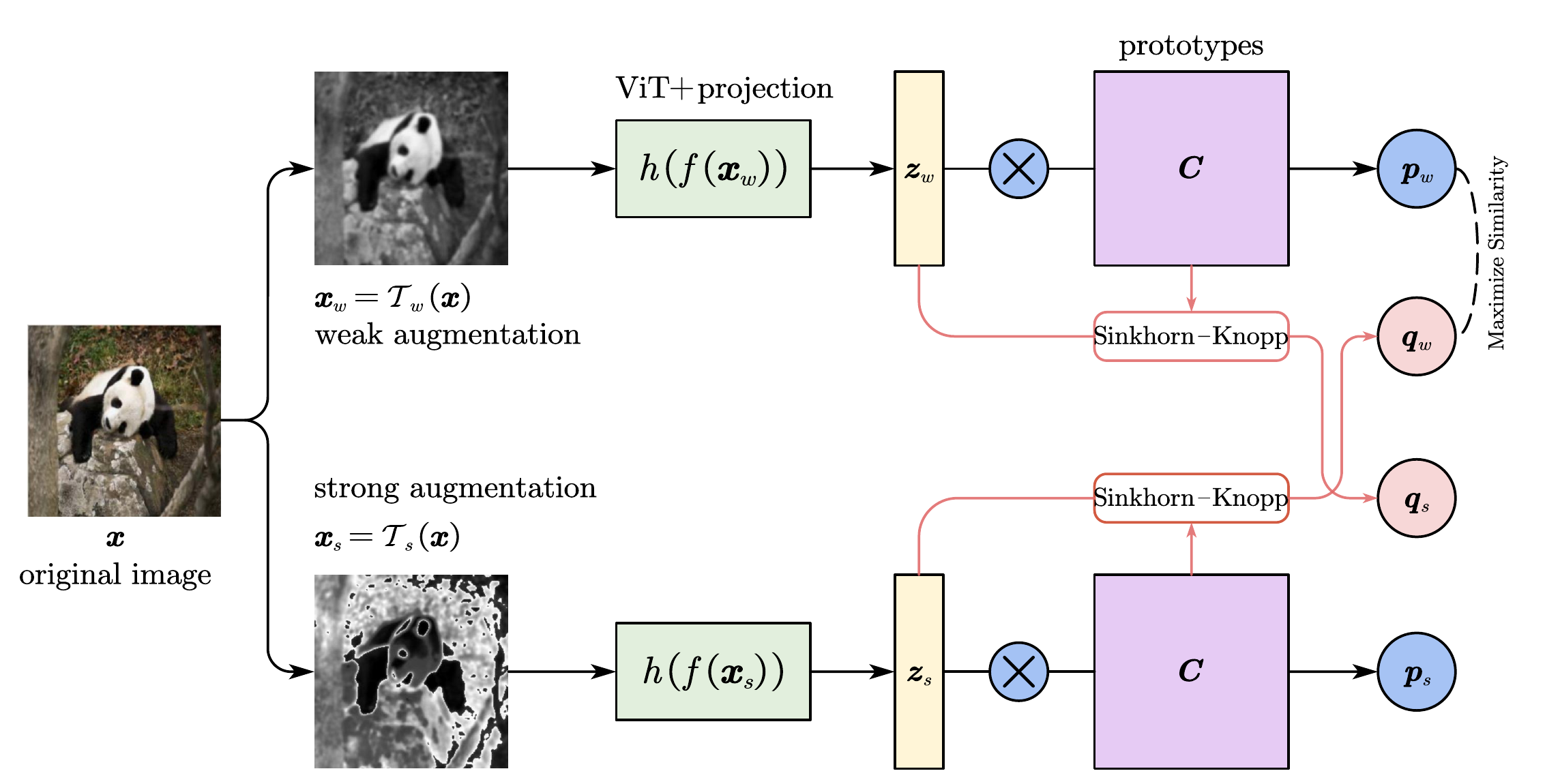}
	\caption{Diagram of Clustering Assignment Consistency. For original image $\bm{x}$, we apply two transformations $\mathcal{T}_w$ and $\mathcal{T}_s$ to obtain two different views $\bm{x}_w$ and $\bm{x}_s$. Then they go through a backbone network, resulting in two embedded representations $\bm{z}_w$ and $\bm{z}_s$. In addition, $\bm{C}$ is the matrix whose columns are the $K$ trainable prototypes vectors, $\{\bm{c}_1,\bm{c}_2,\cdots, \bm{c}_K\}$.
		{\large$\otimes$} denotes the dot products of $\bm{z}_i$ and all prototypes in $\bm{C}$. Then we can compute codes (clustering assignments) $\bm{q}_w$ and $\bm{q}_s$ by matching these representations to the prototypes vectors.}
	\label{fig:pipline}
\end{figure}
\vspace{-15pt}
To tackle the issue of GCD, we leverage co-trained clustering assignment consistency for acquiring discriminative representations of the unlabeled data. We then employ a customized community detection technique to automatically obtain the clustering results and the number of categories. 
Specifically, we use supervised contrastive learning to make maximum use of the supervised information for labeled instances.
As depicted in Figure~\ref{fig:pipline}, we apply a co-training framework to unlabeled data to ensure the consistency between  the clustering assignment and the feature-prototype similarities of the same image.
Following the semi-supervised representation learning, we utilize a modified community detection strategy to allocate category labels for each unlabeled instance, regardless if it is a known one or a novel one, and to determine the number of categories.
\vspace{-15pt}

\subsection{Semi-supervised representation learning}
\vspace{-8pt}
\subsubsection{Supervised contrastive learning.}
We apply supervised contrastive learning to minimize the similarity between embeddings of the same class while maximize the similarity among embeddings from different classes  at the same time. Utilizing label information only in the supervised contrastive learning instead of in a cross-entropy loss results in equal treatment of labelled and unlabeled data. The supervised contrastive component nudges the network towards a semantically meaningful representation to reduce overfitting on the labeled classes.
The supervised contrastive loss, as in \cite{DBLP:conf/nips/KhoslaTWSTIMLK20,DBLP:conf/cvpr/VazeHVZ22}, is defined as,
\begin{equation}\label{eq:sup_con}
	\mathcal{L}_{sup} =
	- \frac{1}{|\mathcal{N}(i)|}  \sum_{p \in \mathcal{N}(i)}\log \frac{\exp \left(sim(\bm{z}_i, \bm{z}_p) / \tau_{sup}\right)}{\sum_{n} \mathds{1}_{[n \ne i]} \exp \left(sim(\bm{z}_i ,\bm{z}_n) / \tau_{sup}\right)},
\end{equation}
where $\bm{z}_i=h(f(\bm{x}_i))$, $f(\cdot)$ is the feature extractor, and $h(\cdot)$ is a multi-layer perceptron (MLP) projection head. $sim(\bm{z}_i, \bm{z}_p)$ denote the cosine similarity between $\bm {z}_i$ and $\bm{z}_p$. $\tau$ is a temperature value.  $\mathcal{N}(i)$ is the indices of other images having the same label as $\bm{x}_i$. $\mathds{1}_{[n\ne i]}$ denotes an indicator function evaluating to $1$ if $n \ne i$. 
\vspace{-15pt}
\subsubsection{Unsupervised contrastive learning.}
We found that unsupervised contrastive learning in \cite{DBLP:conf/cvpr/VazeHVZ22} treats different unlabeled samples from the same semantic category as false negatives, resulting in a class collision problem \cite{DBLP:conf/iccv/Zheng0Y0Z0021}. 
Inspired by this, we introduce a soft approach to mitigate this problem. 
In contrast learning \cite{DBLP:conf/icml/ChenK0H20}, two different augmentations generate two different views,  and this scenario  satisfies the co-training assumption, that is,
\begin{equation}\label{equ:co-asump}
	f_1(\bm{v}_1)=f_2(\bm{v}_2), \forall \bm{x}=(\bm{v}_1,\bm{v}_2) \quad \text{(Co-Training Assumption)},
\end{equation}
where $f(\cdot),f_1(\cdot),f_2(\cdot)$ refer to  different prediction functions. $\bm{v}_1$ and $\bm{v}_2$ are two different views of $\bm{x}$. In this paper, weak augmentation and strong augmentation are utilized to generate two separate views. We rethink both the co-training assumption and the swapped prediction problem discussed in  \cite{DBLP:conf/nips/CaronMMGBJ20}. We consider the probability of feature similarity to the prototypes and promote the minimization of divergence between the soft clustering assignment and the probability in the weakly augmented case. Here, We ignore the case of strong augmentation since the induced distortions could significantly alter the image structures and make it difficult to preserve the identity of the original instances. Additionally, we measure the fit between the feature embeddings and their soft assignments.

Formally, let $\bm{x}_w$ and $\bm{x}_s$ denote the weakly augmented and strongly augmented views of the same image $\bm{x}$, respectively. $\{\bm{c}_1,\bm{c}_2,\cdots,\bm{c}_K\}$ is a set of $K$ learnable prototypes. The feature is projected to the unit sphere, \ie, $\bm{z}_w=\frac{f(\bm{x}_w)}{\|f(\bm{x}_w)\|_2}$. We consider the feature-prototype similarity probability and the clustering assignment as two different predictions as in Eq.~(\ref{equ:co-asump}). Following \cite{DBLP:conf/eccv/QiaoSZWY18}, We use a measure of similarity, the Jensen-Shannon divergence between feature-prototype similarity probability $\bm{p}_w$ and clustering assignment $\bm{q}_w$, the consistency loss is written as 
\begin{equation}
	\mathcal{L}(\bm{z}_w,\bm{q}_w)=H\left(\frac{1}{2}(\bm{p}_w^{(k)}+\bm{q}_w^{(k)})\right)-\frac{1}{2}\left(H(\bm{p}_w^{(k)})+H(\bm{q}_w^{(k)})\right),
\end{equation}
where $H(\bm{p})$ is  the entropy of $\bm{p}$, and
\begin{equation}
	\bm{p}_w^{(k)}=\frac{\exp (sim(\bm{z}_w,\bm{c}_k)/\tau_u)}{\sum_{n} \mathds{1}_{[n \ne i]} \exp \left(sim(\bm{z}_w,\bm{c}_n) / \tau_u\right)}.
\end{equation}
We encourage the consistency of clustering assignments between weakly and strongly augmented views of the same image, and the loss function is written as follows,
\begin{eqnarray}
	\mathcal{L}(\bm{z}_w,\bm{z}_s)&=&\mathcal{L}(\bm{z}_w,\bm{q}_s)+\mathcal{L}(\bm{z}_s, \bm{q}_w)\\
	&=&H(\bm{p}_w,\bm{q}_s)+H(\bm{p}_s, \bm{q}_w)
\end{eqnarray}
where $H(\cdot, \cdot)$ represents the cross entropy loss, 
\begin{equation}
	\bm{p}_s^{(k)}=\frac{\exp (sim(\bm{z}_s,\bm{c}_k)/\tau_u)}{\sum_{n} \mathds{1}_{[n \ne i]} \exp \left(sim(\bm{z}_s,\bm{c}_n) / \tau_u\right)},
\end{equation}
and $sim(\cdot, \cdot)$ denote the cosine similarity as in Eq.~(\ref{eq:sup_con}). Similar to \cite{DBLP:conf/eccv/CaronBJD18,DBLP:conf/nips/CaronMMGBJ20}, we compute clustering assignment matrix $\bm{Q}=[\bm{q}_1,\bm{q}_2,\cdots,\bm{q}_B]$ with $B$ feature embeddings $\bm{Z}=[\bm{z}_1,\bm{z}_2, \cdots \bm{z}_B]$ and the prototypes $\bm{C}=[\bm{c}_1,\bm{c}_2, \cdots, \bm{c}_K]$, 
\begin{eqnarray}\label{equ:Q}
	&\max_{\bm{Q}\in \mathcal{Q}}\textbf{Tr}(\bm{Q}^T\bm{C}^T\bm{Z})+\epsilon H(\bm{Q}),\\
	&\mathcal{Q}=\left\{\bm{Q}\in \mathbb{R}_{+}^{K\times B}|\bm{Q}\bm{1}_B=\frac{1}{K}\bm{1}_K,\bm{Q}^T\bm{1}_K=\frac{1}{B}\bm{1}_B\right\},
\end{eqnarray}
where $\epsilon$ is a weight coefficient. $\textbf{Tr}(\cdot)$ denotes the trace function, and $\mathcal{Q}$ is the transportation polytope. The solution to Eq.~(\ref{equ:Q}) is obtaining using the Sinkhorn-Knopp algorithm \cite{DBLP:conf/nips/Cuturi13}. The pseudo-code for computing $\bm{Q}$ and training loop is in Algorithm~\ref{alg:co-swav}. 

Thus, the overall loss to optimize the model can be formulated as follows,
\begin{equation}
	\mathcal{L}=\mathbb{E}_{(\bm{x},y)\in \mathcal{D}_L}\mathcal{L}_{sup}+\alpha \mathbb{E}_{\bm{x}\in \mathcal{D}}\mathcal{L}(\bm{z}_w,\bm{q}_w) + (1-\alpha)\mathbb{E}_{\bm{x}\in \mathcal{D}}\mathcal{L}(\bm{z}_w,\bm{z}_s),
\end{equation}
where $\alpha$ is a weight coefficient.
\vspace{-15pt}
\SetKwComment{Comment}{/* }{ */}

\begin{algorithm2e}[ht]
	\caption{The torch style pseudo-code for unsupervised contrastive learning.}
	\label{alg:co-swav}
 \KwIn{batch size $B$, prototypes $C$, extractor $f\left(\cdot\right)$, MLP $h\left(\cdot\right)$, temperature $\tau$, dataset $\mathcal{D}_{U}$; }

    \SetKwFunction{FMain}{sinkhorn}
    \SetKwProg{Fn}{Function}{:}{}
    \Fn{\FMain{$scores$, $\epsilon=0.05$, $n_{iters}=3$}}{
        $u^{(0)} =1,  K = \exp(scores / \epsilon) $; 
        
    \For{$i=1$ \KwTo $n_{iters}$}
{   
$v^{(i)}=1 / K^T u^{(i-1)}$;
$u^{(i)}=1 / K^T v^{(i)}$;
}
        \textbf{return} $\text{diag}(u^{(n_{iters})})K \text{diag}(v^{(n_{iters})})$ 
}
\textbf{End Function}

\For {$x \in$ minibatch}{
{$x_w \gets h(f(\text{WeakAugment}(x)))$;}\\
{$x_s \gets h(f(\text{StrongAugment}(x)))$;}\\
{$score_w \gets x_w \times C$;}\\
{$score_s \gets x_s \times C$;}\\
{$p_w = \text{Softmax}(scores_w / \tau)$;}\\
{$p_s = \text{Softmax}(scores_s / \tau)$;}\\
{stop gradients to $score_w$, $score_s$;}\\
{$q_w = \text{sinkhorn}(scores_w)$;}\\
{$q_s = \text{sinkhorn}(scores_s)$;}\\
{compute $l$ by Eq. (10);}\\
{update $h$ and $C$ to minimize $l$;}\\
{$C \gets \text{Normlization}(C)$;}\\
}
\end{algorithm2e}

\subsection{Label assignment with Louvain algorithm}
\vspace{-8pt}
Semi-supervised k-means \cite{DBLP:conf/soda/ArthurV07} is applied in \cite{DBLP:conf/cvpr/VazeHVZ22} for assigning cluster labels. However, this approach necessitates prior knowledge of the number of clusters, a requirement that is not realistic because such information is typically not available beforehand. The learned embeddings can be used to assign cluster labels to unlabeled instances and determine the number of clusters adaptively through a community detection method, Louvain algorithm \cite{Blondel2008FastUO,Dugu2015DirectedL}.
The representation embeddings can be represented as an embedding graph $G=(\mathcal{V},\mathcal{E})$, where $\mathcal{V}=\{\bm{z}_i\}_{i=1}^N$, $N$ is the number of embeddings, and $\mathcal{E}$ consists of edges that connect vertices in $\mathcal{V}$. We define a matrix $\bm{W}$ as the adjacency matrix of $G$, and the entry $\bm{W}_{i,j}$ between  vertices $i$ and $j$ is given by,
\begin{equation}\label{eq:adjecent}
	{\bm{W}}_{ij}= \begin{cases}1, & \text { if } \bm{x}_i,\bm{x}_j \in \mathcal{D}_L, \text{and} ~{y}_{i}={y}_{j}\\ 
		sim(\bm{z}_i,\bm{z}_j), & \text { if } \bm{x}_i ~ \text{or} ~ \bm{x}_j \in \mathcal{D}_U, \text{and} ~\bm{z}_j \in \text{Neighbor}(\bm{z}_i)
		\\
		0, & \text { otherwise }\end{cases},
\end{equation}
where $\text{Neighbor}(\bm{z}_i)=\arg {topM}(\{sim(\bm{z}_i,\bm{z}_j)|\bm{z}_j\in \mathcal{V}\})$ denotes the neighbors of $\bm{z}_i$,\ie, the $M$ embeddings with the greatest similarity to $\bm{z}_i$. In this way, we construct a graph $G$, which represents the possible connection relationships among all instances. 
We can now assign category labels for all instances in the training dataset, either from the known classes or novel ones. Using the Louvain algorithm, we automatically obtain the clustering assignments and the number of categories on the constructed graph.
\vspace{-15pt}



\section{Experiment}
\vspace{-12pt}
\subsection{Datasets}
\vspace{-12pt}
We evaluate our framework on three generic object recognition datasets, namely CIFAR-10~\cite{cifar}, CIFAR-100~\cite{cifar} and ImageNet-100~\cite{DBLP:conf/cvpr/DengDSLL009}. These standard image recognition datasets establish the performance of different methods.
We further evaluate our method on Semantic Shift Benchmark (SSB) \cite{DBLP:conf/iclr/Vaze0VZ22}, includeing  CUB-200~\cite{dataset-cub}, Stanford Cars~\cite{scars} and Herbarium19~\cite{DBLP:journals/corr/abs-1906-05372}. The dataset splits are described in \ref{tab:datasets}. We follow \cite{DBLP:conf/cvpr/VazeHVZ22} sample a subset of half the classes as `'Old" categories. $50\%$ of instances of each labeled class are drawn to form the labeled set, and all the remaining data constitute the unlabeled set. For evaluation, we measure the clustering accuracy by comparing the predicted label assignment with the ground truth, following the protocol in  \cite{DBLP:conf/cvpr/VazeHVZ22}. The accuracy scores for \texttt{All}, \texttt{Known}, and \texttt{Novel} categories are reported.
\begin{table*}[ht]
	\centering
	\caption{Dataset splits in the experiments.}
	\label{tab:datasets}
	\resizebox{\linewidth}{!}{
		\begin{tabular}{cc|c|c|c|c|c|c}
			\toprule
			\multicolumn{2}{c|}{\textbf{Dataset}}    & \textbf{CIFAR10} & \textbf{CIFAR100}  & \textbf{ImageNet-100}  & \textbf{CUB-200} & \textbf{Stanford Cars} & \textbf{Herbarium19}
			\\
			\midrule
			\multirow{2}{*}{Labelled}   & Classes & 5       &  80      & 50           & 100     & 98    & 341           \\
			& Images  & 12.5k   & 20k & 31.9k  & 1498    & 2000  & 8.9k      \\
			\hline
			\multirow{2}{*}{Unlabelled} & Classes & 10      &  100       & 100     & 200     & 196   & 683        \\
			& Images  & 37.5k   & 30k & 95.3k  & 4496    & 6144  & 25.4k     \\
			\bottomrule
	\end{tabular}}
\end{table*}

\subsection{Implementation Details.}
\vspace{-5pt}
We follow the implementations and learn schedules in \cite{DBLP:conf/cvpr/VazeHVZ22} as far as possible.  Specifically, we take the ViT-B-16 pre-trained by DINO \cite{DBLP:conf/iccv/CaronTMJMBJ21} on ImageNet \cite{DBLP:journals/cacm/KrizhevskySH17} as our backbone model and we use the \texttt{[CLS]} token as the feature representation. 
We train semi-supervised contrastive learning with $\tau_{sup}=0.07$, $\tau_u=0.05$ and $\alpha=0.3$. For image augmentation, we use ResizedCrop, ColorJitter, Grayscale, HorizontalFlip as the weak image augmentation. The strong transformation strategy is composed of five randomly selected from RandAugment. For model optimization, we use the AdamW provided by \cite{DBLP:conf/iclr/LoshchilovH19}. The initial learning rate is 0.1. The batch size are chosen based on available GPU memory. All the experiments are conducted on a single RTX-2080 and averaged over 5 different seeds. 
\vspace{-15pt}
\subsection{Compare with State-of-the-Arts}
\vspace{-5pt}
We summarize the baselines compared in our experiments, including $k$-means~\cite{DBLP:conf/soda/ArthurV07}, RankStats+~\cite{DBLP:conf/iclr/HanREVZ20}, UNO+~\cite{DBLP:conf/iccv/FiniSLZN021} and GCD~\cite{DBLP:conf/cvpr/VazeHVZ22}. 
\vspace{-15pt}
	\begin{table}[ht]
		\centering
        \caption{\textbf{Results on three generic datasets.}  Accuracy scores are reported. \dag denotes adapted methods.}
		\label{tab:main-gen}
			\begin{tabular}{c||ccc|ccc|ccc}
				\toprule
				& \multicolumn{3}{c|}{\textbf{CIFAR-10}} & \multicolumn{3}{c|}{\textbf{CIFAR-100}} & \multicolumn{3}{c}{\textbf{ImageNet-100}} \\
				\textbf{Method} & All & Known & Novel & All & Known & Novel & All & Known & Novel \\
				\midrule
				KMeans~\cite{DBLP:conf/soda/ArthurV07} & 83.6 & 85.7 & 82.5 & 52.0 & 52.2 & 50.8 & 72.7 & 75.5 & 71.3 \\
				RankStats\dag ~\cite{DBLP:conf/iclr/HanREVZ20} &  46.8 & 19.2 & 60.5 & 58.2 & 77.6 & 19.3 & 37.1 & 61.6 & 24.8 \\
				UNO\dag~\cite{DBLP:conf/iccv/FiniSLZN021}& 68.6 & \textbf{98.3} & 53.8 & 69.5 & 80.6 & 47.2 & 70.3 & \textbf{95.0} & 57.9 \\
				GCD~\cite{DBLP:conf/cvpr/VazeHVZ22} & 91.5 & 97.9 & 88.2 & 73.0 & 76.2 & 66.5 & 74.1 & 89.8 & 66.3 \\
				\midrule
				\textbf{Ours} & \textbf{92.3} & 91.4 & \textbf{94.4} & \textbf{78.5} & \textbf{81.4} & \textbf{75.6} & \textbf{81.1} & 80.3 & \textbf{81.8} \\
				
				\bottomrule
			\end{tabular}
	\end{table}

\vspace{-15pt}
\textbf{Evaluation on generic datasets.} 
The results on generic benchmarks are shown in Table~\ref{tab:main-gen}. As we can see that our method achieves state-of-the art performance on \texttt{All} and \texttt{Novel} tested on all generic datasets, especially on ImageNet-100. Our method also achieves comparable results with other methods on \texttt{Known}. Specifically, for the \texttt{All} classes, our method beats the GCD method by 0.8\%, 5.5\%, and 7.0\% on CIFAR-10, CIFAR-100, and ImageNet-100, respectively. For the \texttt{Novel} class, it is 6.2\% higher on CIFAR-10, 9.1\% higher on CIFAR-100, and 15.5\% higher on ImageNet-100. These results experimentally show that our method learns a more compact representation on the unlabeled dataset. In addition, UNO+ uses a linear classifier, which shows strong accuracy on the \texttt{Known} classes, but leads to poor performance on the \texttt{Novel} classes.

\begin{table}[ht]
\centering
\caption{\textbf{Results on three fine-grained datasets.} Accuracy scores are reported. \dag denotes adapted methods.}
\label{tab:fine_grained}
\begin{tabular}{c||ccc|ccc|ccc}
\toprule
 & \multicolumn{3}{c|}{\textbf{CUB-200}} & \multicolumn{3}{c|}{\textbf{Stanford-Cars}} & \multicolumn{3}{c}{\textbf{Herbarium19}} \\
\textbf{Method }      & All      & Known    & Novel        & All       & Known      & Novel       & All       & Known       & Novel   \\
\midrule
$k$-means~\cite{DBLP:conf/soda/ArthurV07}
& 34.3
& 38.9
& 32.1
& 12.8
& 10.6
& 13.8
& 12.9
& 12.9
& 12.8
\\
RankStats+~\cite{DBLP:conf/iclr/HanREVZ20}
& 33.3
& 51.6
& 24.2
& 28.3
& 61.8
& 12.1
& 27.9
& \textbf{55.8}
& 12.8
\\
UNO+ ~\cite{DBLP:conf/iccv/FiniSLZN021}
& 35.1
& 49.0
& 28.1
& 35.5
& 70.5
& 18.6
& 28.3
& 53.7
& 14.7
\\
GCD~\cite{DBLP:conf/cvpr/VazeHVZ22}
& 51.3
& 56.6
& \textbf{48.7}
& 39.0
& 57.6
& 29.9
& 35.4
& 51.0
& 27.0
\\
\midrule
\textbf{Ours}
& \textbf{58.0}
& \textbf{65.0}
& 43.9
& \textbf{47.6}
& \textbf{70.6}
& \textbf{33.8}
& \textbf{36.3}
& 53.1
& \textbf{30.7}
\\
\bottomrule
\end{tabular}
\end{table}

\vspace{-15pt}
\textbf{Evaluation on fine-grained datasets.}
We report the results on three fine-grained datasets (as in \cite{DBLP:conf/cvpr/VazeHVZ22} in Table~\ref{tab:fine_grained}. Our method shows optimal performance on the \texttt{All} classes for the  three datasets tested, and achieves comparable results on the \texttt{Known} and \texttt{Novel} classes, demonstrating the effectiveness of our method for fine-grained category discovery. Specifically, on the CUB-200, Stanford-Scars, and Herbarium19 datasets, our method achieves 6.7\%, 8.6\%, and 0.9\% improvement over the state-of-the-art method on the \texttt{All} classes, respectively. For the \texttt{Novel} classes, our method outperforms GCD by 3.9\% and 3.7\% on Stanford-cars and Herbarium19, respectively. Meanwhile, we find that due to the low variability between fine-grained datasets, which makes it more difficult to discover novel classes, the precision on the \texttt{Novel} classes is generally low in terms of results.
\vspace{-8pt}

\subsection{Ablation Study}
\vspace{-8pt}
\textbf{Effectiveness of each component.} We conduct extensive ablation experiments, and perform four experiments on the CIFAR-100 and CUB-200 datasets. Table~\ref{tab:ablation_loss}  shows the contribution of introducing different components on the objective  loss function, including $\mathcal{L}_{sup}$, $\mathcal{L}(\bm{z}_w,\bm{q}_w)$ and $\mathcal{L}(\bm{z}_w, \bm{z}_s)$. We can observe from the results that all components contribute significantly to our proposed approach: according to the results of experiments (1) and (2), we find that $\mathcal{L}(\bm{z}_w,\bm{z}_s)$ is the most important component, which proves that the co-training consistency can close the intra-class embeddings and push away the inter-class boundaries. Moreover, compared to the original DINO features, the features of our approach show more favorable clustering results on CUB-200 dataset. Comparing experiments (3) and (4), we can find that supervised contrastive learning can further improve the performance on known novel categories, demonstrating the importance of supervised information. 

\begin{table}[ht]
	\small
	\centering
	\caption{Ablation study on the components of the loss function.}
	\label{tab:ablation_loss}
	\begin{tabular}{c|ccc|ccc|ccc}
		\toprule
		\multirow{2}{*}{\textbf{Index}} & \multicolumn{3}{c|}{\textbf{Component}}                                                                                  & \multicolumn{3}{c|}{\textbf{CIFAR100}}                                  & \multicolumn{3}{c}{\textbf{CUB-200}}                                   \\ \cline{2-10} 
		& \multicolumn{1}{c}{$\mathcal{L}_{sup}$}            & \multicolumn{1}{c}{$\mathcal{L}(\bm{z}_w,\bm{q}_w)$}          &  $\mathcal{L}(\bm{z}_w,\bm{z}_s)$         & \multicolumn{1}{c}{All}  & \multicolumn{1}{c}{Known} & Novel & \multicolumn{1}{c}{All}  & \multicolumn{1}{c}{Known} & Novel \\ 
		\midrule
		(1)                    & \multicolumn{1}{c}{\xmark} & \multicolumn{1}{c}{\xmark} & \xmark & \multicolumn{1}{c}{34.9} & \multicolumn{1}{c}{36.1}  & 33.6  & \multicolumn{1}{c}{15.0} & \multicolumn{1}{c}{13.5}  & 21.6  \\ 
		(2)                    & \multicolumn{1}{c}{\xmark} & \multicolumn{1}{c}{\xmark} & \cmark & \multicolumn{1}{c}{69.7} & \multicolumn{1}{c}{71.4}  & 68.0  & \multicolumn{1}{c}{56.7} & \multicolumn{1}{c}{62.5}  & 40.8  \\ 
		(3)                    & \multicolumn{1}{c}{\xmark} & \multicolumn{1}{c}{\cmark} & \cmark & \multicolumn{1}{c}{75.7} & \multicolumn{1}{c}{79.2}  & 72.5  & \multicolumn{1}{c}{57.5} & \multicolumn{1}{c}{63.9}  & 42.8  \\ 
		(4)                    & \multicolumn{1}{c}{\cmark} & \multicolumn{1}{c}{\cmark} & \cmark & \multicolumn{1}{c}{78.5} & \multicolumn{1}{c}{81.4}  & 75.5  & \multicolumn{1}{c}{58.0} & \multicolumn{1}{c}{65.0}  & 43.9  \\ \bottomrule
	\end{tabular}
\end{table}

\begin{table}[ht]
	\centering
	\caption{Ablation study on neighborhood size.}
	\label{tab:ablation_M}
	\begin{tabular}{c|ccc|ccc}
		\toprule
		\multirow{2}{*}{M} & \multicolumn{3}{c|}{\textbf{CIFAR-100}} & \multicolumn{3}{c}{\textbf{CUB-200}} \\
		& All     & Known    & Novel    & All     & Known   & Novel   \\
		\midrule
		5                  & 78.5    & 81.4     & 75.5     & 58.0    & 65.0    & 43.9    \\
		10                 & 69.6    & 70.4     & 68.7     & 53.5    & 60.8    & 35.1    \\
		15                 & 64.9    & 65.0     & 64.8     & 46.8    & 54.5    & 26.8    \\
		20                 & 63.5    & 62.3     & 64.8     & 44.5    & 49.8    & 33.7    \\
		25                 & 59.7    & 60.3     & 59.1     & 39.7    & 44.6    & 30.5    \\
		30                 & 59.8    & 60.4     & 59.2     & 27.3    & 29.3    & 29.7   \\
		\bottomrule
	\end{tabular}
\end{table}
\textbf{Effectiveness of the neighborhood size $M$.} Table~\ref{tab:ablation_M} illustrates the effect of the neighborhood size for vertex $\bm{z}_i \in G(\mathcal{V},\mathcal{E})$ on the final clustering results. We select $K=5,10,15,20,25,30$ for the ablation experiments on both CIFAR-100 and CUB-200 datasets. We find that the final performance varies greatly depending on the number of neighbors. When $K=5$, it can reach the optimum in all \texttt{All}, \texttt{Known} and \texttt{Novel} classes. We conjecture that too many links will negatively affect the community detection.

\begin{figure}[ht]
	\begin{minipage}{0.32\linewidth}
		\vspace{3pt}
		\centerline{\includegraphics[width=\textwidth,trim=50 50 50 50,clip]{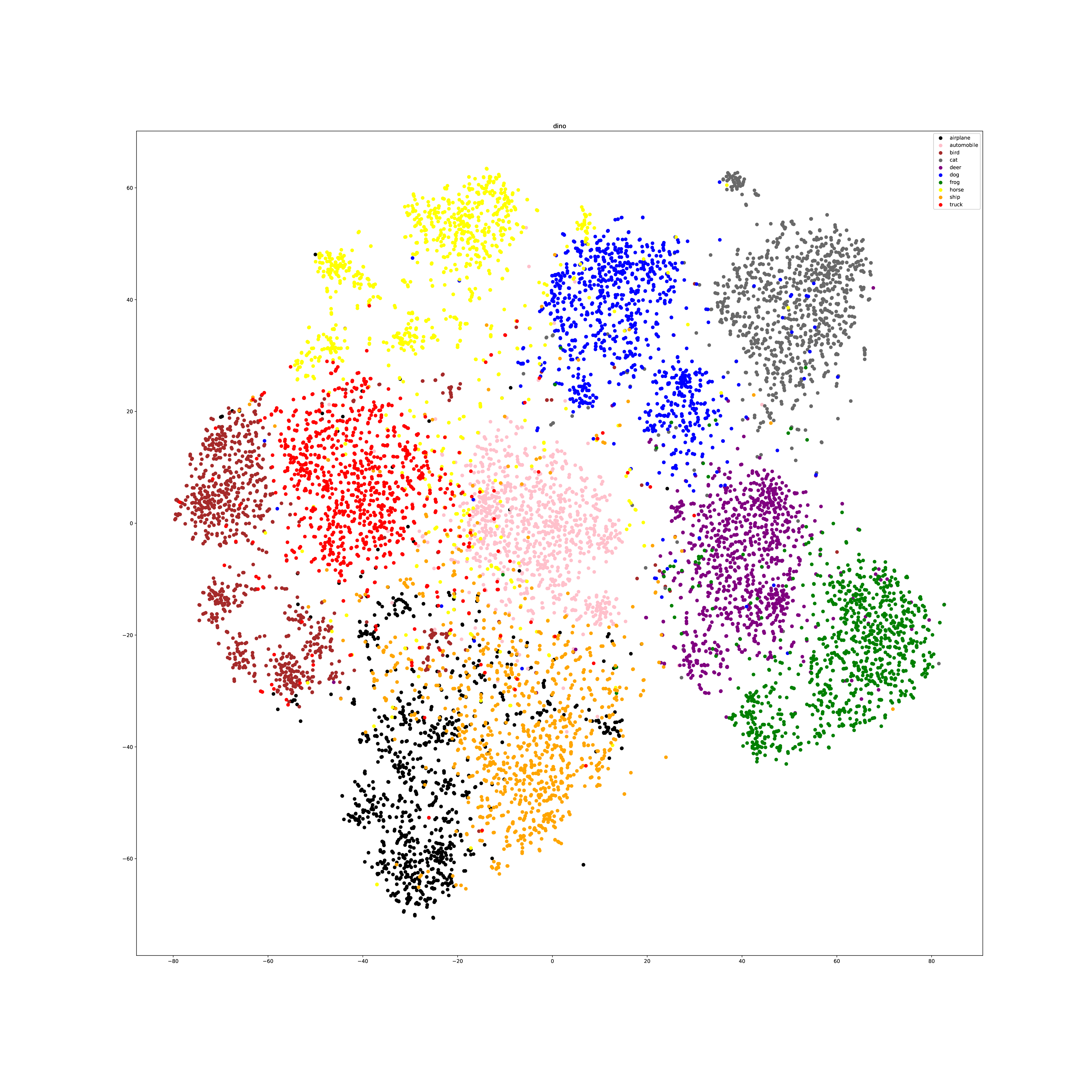}}
		\centerline{(a) DINO}
	\end{minipage}
	\begin{minipage}{0.32\linewidth}
		\vspace{3pt}
		\centerline{\includegraphics[width=\textwidth, trim=50 50 50 50,clip]{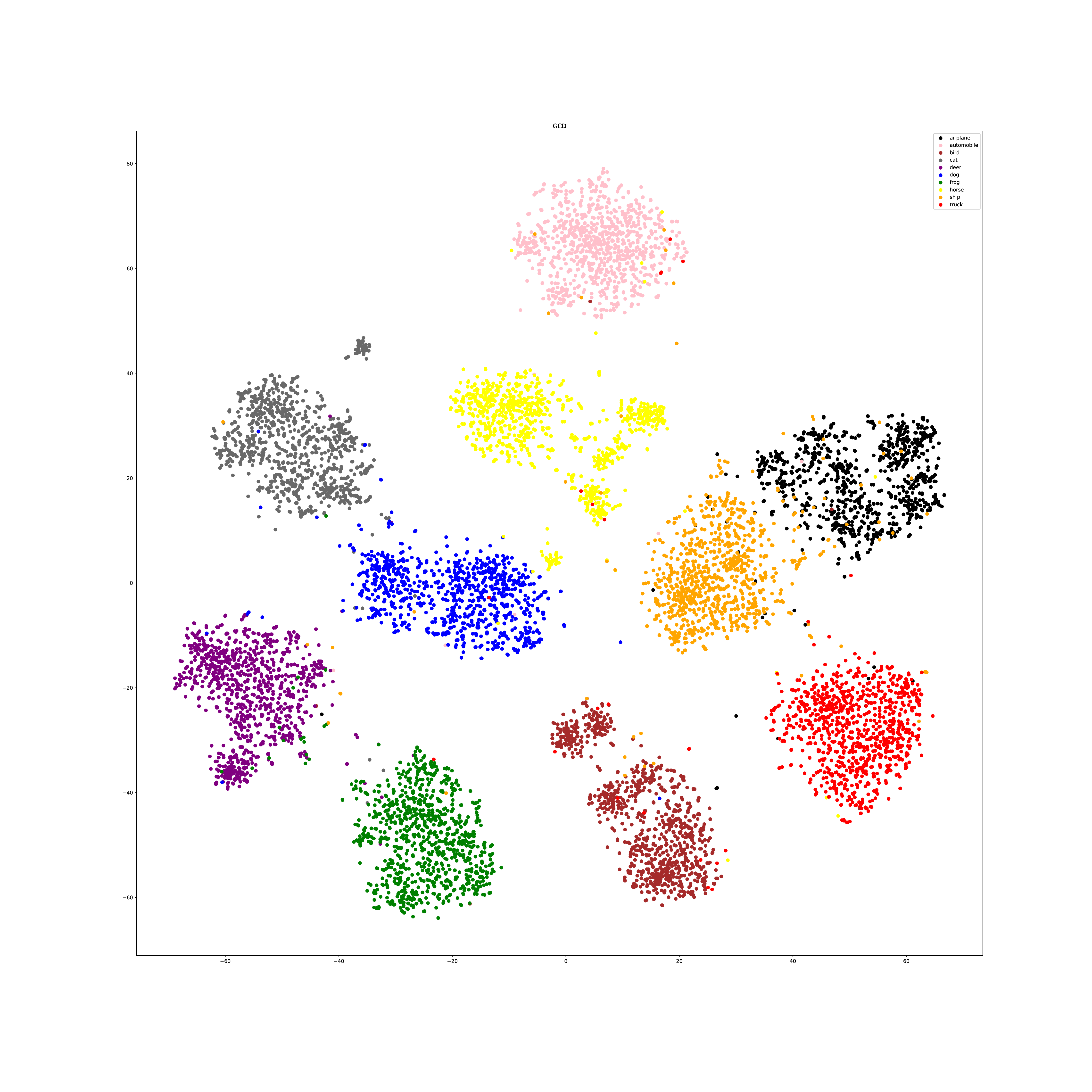}}
		\centerline{(b) GCD}
	\end{minipage}
	\begin{minipage}{0.32\linewidth}
		\vspace{3pt}
		\centerline{\includegraphics[width=\textwidth,trim=50 50 50 50,clip]{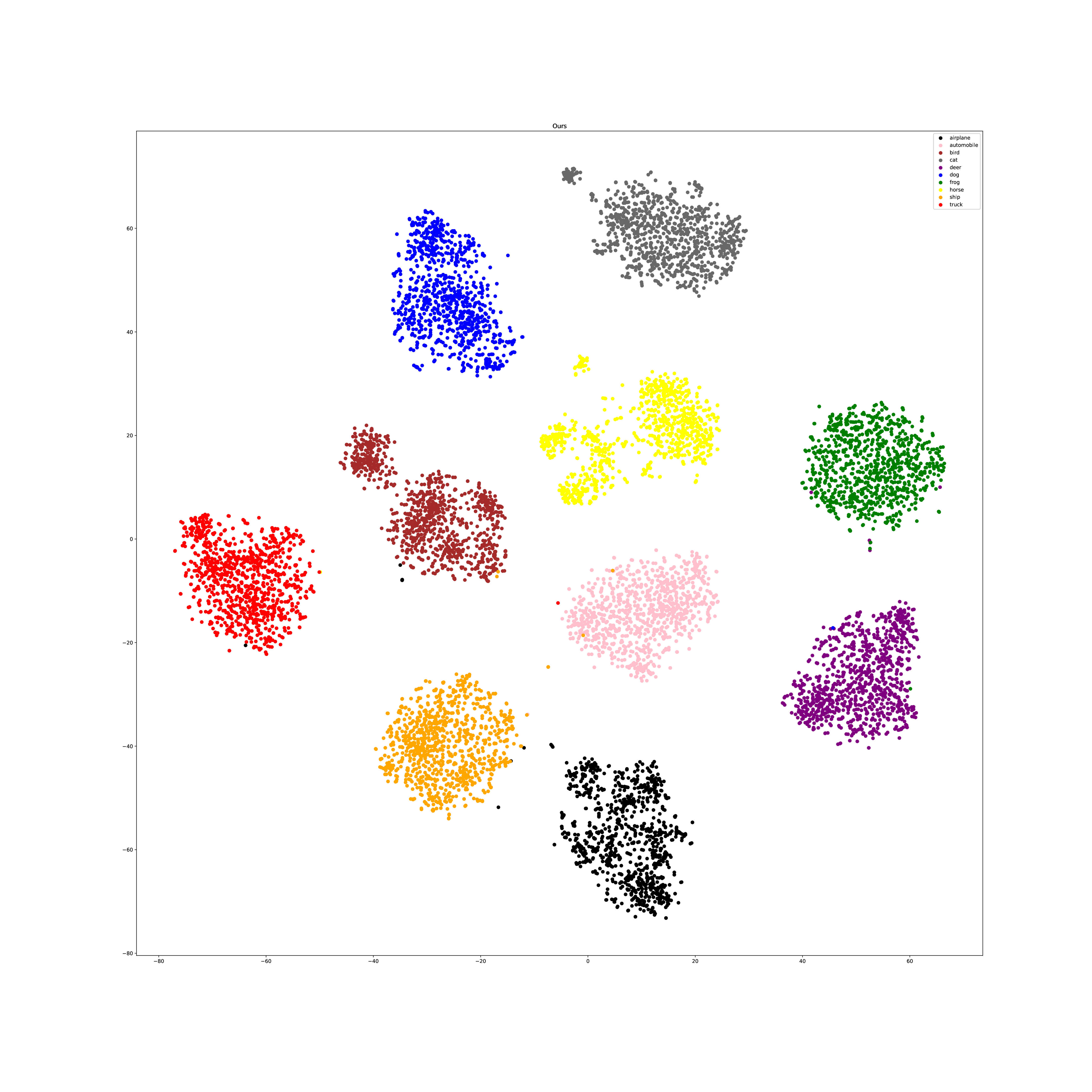}}
		\centerline{(c) Ours}
	\end{minipage}
	\caption{T-SNE visualization of the embeddings on CIFAR-10. The embedding clustering shows that our proposed method encourages the expansion of the distance between different clusters.}
	\label{fig:t-sne}
\end{figure}
\vspace{-1pt}
\textbf{Visualization.}
In order to explore the clustering features on different methods more intuitively, we further visualized the features extracted by DINO, GCD and our method using T-SNE \cite{DBLP:journals/jmlr/Maaten14} on CIFAR-10. As shown in Figure~\ref{fig:t-sne}, compared with DINO and GCD, our method obtains clearer boundaries between different groups, and furthermore obtains more compact clusters.

\vspace{-10pt}
\section{Conclusion}
\vspace{-10pt}
In this paper, we propose a co-training strategy for GCD. To implement this idea, we introduce a clustering assignment consistency framework that explores discriminative representations alternately. Additionally, we propose a community detection method to address the semi-supervised clustering problem in GCD. Experimental results demonstrate that our method achieves state-of-the-art performance in both generic and fine-grained tasks.
\vspace{-15pt}
\subsubsection{Acknowledgements} Please place your acknowledgments at
the end of the paper, preceded by an unnumbered run-in heading (i.e.
3rd-level heading).

%
%
%
%
\vspace{-15pt}
\bibliographystyle{splncs04}
\bibliography{./refer.bib}

\begin{thebibliography}{10}
\providecommand{\url}[1]{\texttt{#1}}
\providecommand{\urlprefix}{URL }
\providecommand{\doi}[1]{https://doi.org/#1}

\bibitem{DBLP:journals/entropy/Albelwi22}
Albelwi, S.: Survey on self-supervised learning: Auxiliary pretext tasks and
  contrastive learning methods in imaging. Entropy  \textbf{24}(4), ~551 (2022)

\bibitem{DBLP:journals/corr/abs-2211-15115}
An, W., Tian, F., Zheng, Q., Ding, W., Wang, Q., Chen, P.: Generalized category
  discovery with decoupled prototypical network. CoRR  \textbf{abs/2211.15115}
  (2022)

\bibitem{DBLP:conf/soda/ArthurV07}
Arthur, D., Vassilvitskii, S.: k-means++: the advantages of careful seeding.
  In: {SODA}. pp. 1027--1035. {SIAM} (2007)

\bibitem{DBLP:conf/nips/BerthelotCGPOR19}
Berthelot, D., Carlini, N., Goodfellow, I.J., Papernot, N., Oliver, A., Raffel,
  C.: Mixmatch: {A} holistic approach to semi-supervised learning. In: NeurIPS.
  pp. 5050--5060 (2019)

\bibitem{Blondel2008FastUO}
Blondel, V.D., Guillaume, J.L., Lambiotte, R., Lefebvre, E.: Fast unfolding of
  communities in large networks. Journal of Statistical Mechanics: Theory and
  Experiment  \textbf{2008},  P10008 (2008)

\bibitem{DBLP:conf/iclr/CaoBL22}
Cao, K., Brbic, M., Leskovec, J.: Open-world semi-supervised learning. In:
  {ICLR}. OpenReview.net (2022)

\bibitem{DBLP:conf/eccv/CaronBJD18}
Caron, M., Bojanowski, P., Joulin, A., Douze, M.: Deep clustering for
  unsupervised learning of visual features. In: {ECCV} {(14)}. Lecture Notes in
  Computer Science, vol. 11218, pp. 139--156. Springer (2018)

\bibitem{DBLP:conf/nips/CaronMMGBJ20}
Caron, M., Misra, I., Mairal, J., Goyal, P., Bojanowski, P., Joulin, A.:
  Unsupervised learning of visual features by contrasting cluster assignments.
  In: NeurIPS (2020)

\bibitem{DBLP:conf/iccv/CaronTMJMBJ21}
Caron, M., Touvron, H., Misra, I., J{\'{e}}gou, H., Mairal, J., Bojanowski, P.,
  Joulin, A.: Emerging properties in self-supervised vision transformers. In:
  {ICCV}. pp. 9630--9640. {IEEE} (2021)

\bibitem{DBLP:conf/icml/ChenK0H20}
Chen, T., Kornblith, S., Norouzi, M., Hinton, G.E.: A simple framework for
  contrastive learning of visual representations. In: {ICML}. Proceedings of
  Machine Learning Research, vol.~119, pp. 1597--1607. {PMLR} (2020)

\bibitem{DBLP:conf/cvpr/ChenH21}
Chen, X., He, K.: Exploring simple siamese representation learning. In: {CVPR}.
  pp. 15750--15758. Computer Vision Foundation / {IEEE} (2021)

\bibitem{DBLP:conf/iclr/ChiLYLL00ZS22}
Chi, H., Liu, F., Yang, W., Lan, L., Liu, T., Han, B., Niu, G., Zhou, M.,
  Sugiyama, M.: Meta discovery: Learning to discover novel classes given very
  limited data. In: {ICLR}. OpenReview.net (2022)

\bibitem{DBLP:conf/nips/Cuturi13}
Cuturi, M.: Sinkhorn distances: Lightspeed computation of optimal transport.
  In: {NIPS}. pp. 2292--2300 (2013)

\bibitem{DBLP:conf/cvpr/DengDSLL009}
Deng, J., Dong, W., Socher, R., Li, L., Li, K., Fei{-}Fei, L.: Imagenet: {A}
  large-scale hierarchical image database. In: {CVPR}. pp. 248--255. {IEEE}
  Computer Society (2009)

\bibitem{DBLP:conf/iclr/DhillonCRS20}
Dhillon, G.S., Chaudhari, P., Ravichandran, A., Soatto, S.: A baseline for
  few-shot image classification. In: {ICLR}. OpenReview.net (2020)

\bibitem{DBLP:conf/iccv/DoerschGE15}
Doersch, C., Gupta, A., Efros, A.A.: Unsupervised visual representation
  learning by context prediction. In: {ICCV}. pp. 1422--1430. {IEEE} Computer
  Society (2015)

\bibitem{DBLP:conf/iclr/DosovitskiyB0WZ21}
Dosovitskiy, A., Beyer, L., Kolesnikov, A., Weissenborn, D., Zhai, X.,
  Unterthiner, T., Dehghani, M., Minderer, M., Heigold, G., Gelly, S.,
  Uszkoreit, J., Houlsby, N.: An image is worth 16x16 words: Transformers for
  image recognition at scale. In: {ICLR}. OpenReview.net (2021)

\bibitem{DBLP:journals/pami/DosovitskiyFSRB16}
Dosovitskiy, A., Fischer, P., Springenberg, J.T., Riedmiller, M.A., Brox, T.:
  Discriminative unsupervised feature learning with exemplar convolutional
  neural networks. {IEEE} Trans. Pattern Anal. Mach. Intell.  \textbf{38}(9),
  1734--1747 (2016)

\bibitem{Dugu2015DirectedL}
Dugu{\'e}, N., Perez, A.: Directed louvain : maximizing modularity in directed
  networks (2015)

\bibitem{DBLP:conf/bmvc/FeiZYZ22}
Fei, Y., Zhao, Z., Yang, S., Zhao, B.: Xcon: Learning with experts for
  fine-grained category discovery. In: {BMVC}. p.~96. {BMVA} Press (2022)

\bibitem{DBLP:conf/iccv/FiniSLZN021}
Fini, E., Sangineto, E., Lathuili{\`{e}}re, S., Zhong, Z., Nabi, M., Ricci, E.:
  A unified objective for novel class discovery. In: {ICCV}. pp. 9264--9272.
  {IEEE} (2021)

\bibitem{DBLP:conf/icml/FinnAL17}
Finn, C., Abbeel, P., Levine, S.: Model-agnostic meta-learning for fast
  adaptation of deep networks. In: {ICML}. Proceedings of Machine Learning
  Research, vol.~70, pp. 1126--1135. {PMLR} (2017)

\bibitem{DBLP:conf/iclr/GidarisSK18}
Gidaris, S., Singh, P., Komodakis, N.: Unsupervised representation learning by
  predicting image rotations. In: {ICLR} (Poster). OpenReview.net (2018)

\bibitem{DBLP:conf/nips/GrillSATRBDPGAP20}
Grill, J., Strub, F., Altch{\'{e}}, F., Tallec, C., Richemond, P.H.,
  Buchatskaya, E., Doersch, C., Pires, B.{\'{A}}., Guo, Z., Azar, M.G., Piot,
  B., Kavukcuoglu, K., Munos, R., Valko, M.: Bootstrap your own latent - {A}
  new approach to self-supervised learning. In: NeurIPS (2020)

\bibitem{DBLP:conf/iclr/HanREVZ20}
Han, K., Rebuffi, S., Ehrhardt, S., Vedaldi, A., Zisserman, A.: Automatically
  discovering and learning new visual categories with ranking statistics. In:
  {ICLR}. OpenReview.net (2020)

\bibitem{DBLP:journals/pami/HanREVZ22}
Han, K., Rebuffi, S., Ehrhardt, S., Vedaldi, A., Zisserman, A.: Autonovel:
  Automatically discovering and learning novel visual categories. {IEEE} Trans.
  Pattern Anal. Mach. Intell.  \textbf{44}(10),  6767--6781 (2022)

\bibitem{DBLP:conf/iccv/HanVZ19}
Han, K., Vedaldi, A., Zisserman, A.: Learning to discover novel visual
  categories via deep transfer clustering. In: {ICCV}. pp. 8400--8408. {IEEE}
  (2019)

\bibitem{DBLP:conf/cvpr/He0WXG20}
He, K., Fan, H., Wu, Y., Xie, S., Girshick, R.B.: Momentum contrast for
  unsupervised visual representation learning. In: {CVPR}. pp. 9726--9735.
  Computer Vision Foundation / {IEEE} (2020)

\bibitem{DBLP:conf/cvpr/HeZRS16}
He, K., Zhang, X., Ren, S., Sun, J.: Deep residual learning for image
  recognition. In: {CVPR}. pp. 770--778. {IEEE} Computer Society (2016)

\bibitem{DBLP:journals/natmi/HolmbergKMSHKAS20}
Holmberg, O.G., K{\"{o}}hler, N.D., Martins, T., Siedlecki, J., Herold, T.,
  Keidel, L., Asani, B., Schiefelbein, J., Priglinger, S., Kortuem, K.U.,
  Theis, F.J.: Self-supervised retinal thickness prediction enables deep
  learning from unlabelled data to boost classification of diabetic
  retinopathy. Nat. Mach. Intell.  \textbf{2}(11),  719--726 (2020)

\bibitem{DBLP:conf/iclr/HsuLK18}
Hsu, Y., Lv, Z., Kira, Z.: Learning to cluster in order to transfer across
  domains and tasks. In: {ICLR} (Poster). OpenReview.net (2018)

\bibitem{DBLP:conf/iclr/HsuLSOK19}
Hsu, Y., Lv, Z., Schlosser, J., Odom, P., Kira, Z.: Multi-class classification
  without multi-class labels. In: {ICLR} (Poster). OpenReview.net (2019)

\bibitem{DBLP:conf/nips/KhoslaTWSTIMLK20}
Khosla, P., Teterwak, P., Wang, C., Sarna, A., Tian, Y., Isola, P., Maschinot,
  A., Liu, C., Krishnan, D.: Supervised contrastive learning. In: NeurIPS
  (2020)

\bibitem{scars}
Krause, J., Stark, M., Deng, J., Fei-Fei, L.: 3d object representations for
  fine-grained categorization. In: Proceedings of the IEEE international
  conference on computer vision workshops. pp. 554--561 (2013)

\bibitem{cifar}
Krizhevsky, A., Hinton, G., et~al.: Learning multiple layers of features from
  tiny images  (2009)

\bibitem{DBLP:journals/cacm/KrizhevskySH17}
Krizhevsky, A., Sutskever, I., Hinton, G.E.: Imagenet classification with deep
  convolutional neural networks. Commun. {ACM}  \textbf{60}(6),  84--90 (2017)

\bibitem{DBLP:conf/iclr/LoshchilovH19}
Loshchilov, I., Hutter, F.: Decoupled weight decay regularization. In: {ICLR}
  (Poster). OpenReview.net (2019)

\bibitem{DBLP:journals/jmlr/Maaten14}
van~der Maaten, L.: Accelerating t-sne using tree-based algorithms. J. Mach.
  Learn. Res.  \textbf{15}(1),  3221--3245 (2014)

\bibitem{DBLP:conf/iclr/MitrovicMWBB21}
Mitrovic, J., McWilliams, B., Walker, J.C., Buesing, L.H., Blundell, C.:
  Representation learning via invariant causal mechanisms. In: {ICLR}.
  OpenReview.net (2021)

\bibitem{DBLP:conf/eccv/NorooziF16}
Noroozi, M., Favaro, P.: Unsupervised learning of visual representations by
  solving jigsaw puzzles. In: {ECCV} {(6)}. Lecture Notes in Computer Science,
  vol.~9910, pp. 69--84. Springer (2016)

\bibitem{DBLP:conf/cvpr/PathakKDDE16}
Pathak, D., Kr{\"{a}}henb{\"{u}}hl, P., Donahue, J., Darrell, T., Efros, A.A.:
  Context encoders: Feature learning by inpainting. In: {CVPR}. pp. 2536--2544.
  {IEEE} Computer Society (2016)

\bibitem{DBLP:conf/eccv/QiaoSZWY18}
Qiao, S., Shen, W., Zhang, Z., Wang, B., Yuille, A.L.: Deep co-training for
  semi-supervised image recognition. In: {ECCV} {(15)}. Lecture Notes in
  Computer Science, vol. 11219, pp. 142--159. Springer (2018)

\bibitem{DBLP:conf/cvpr/RohSKK21}
Roh, B., Shin, W., Kim, I., Kim, S.: Spatially consistent representation
  learning. In: {CVPR}. pp. 1144--1153. Computer Vision Foundation / {IEEE}
  (2021)

\bibitem{DBLP:journals/corr/abs-1906-05372}
Tan, K.C., Liu, Y., Ambrose, B., Tulig, M., Belongie, S.J.: The herbarium
  challenge 2019 dataset. CoRR  \textbf{abs/1906.05372} (2019)

\bibitem{DBLP:journals/corr/abs-2302-12028}
Troisemaine, C., Lemaire, V., Gosselin, S., Reiffers{-}Masson, A.,
  Flocon{-}Cholet, J., Vaton, S.: Novel class discovery: an introduction and
  key concepts. CoRR  \textbf{abs/2302.12028} (2023)

\bibitem{DBLP:conf/cvpr/VazeHVZ22}
Vaze, S., Han, K., Vedaldi, A., Zisserman, A.: Generalized category discovery.
  In: {CVPR}. pp. 7482--7491. {IEEE} (2022)

\bibitem{DBLP:conf/iclr/Vaze0VZ22}
Vaze, S., Han, K., Vedaldi, A., Zisserman, A.: Open-set recognition: {A} good
  closed-set classifier is all you need. In: {ICLR}. OpenReview.net (2022)

\bibitem{dataset-cub}
Wah, C., Branson, S., Welinder, P., Perona, P., Belongie, S.: The caltech-ucsd
  birds-200-2011 dataset  (2011)

\bibitem{DBLP:journals/corr/abs-2205-07246}
Wang, Y., Chen, H., Heng, Q., Hou, W., Fan, Y., Wu, Z., Savvides, M.,
  Shinozaki, T., Raj, B., Schiele, B.: Freematch: Self-adaptive thresholding
  for semi-supervised learning. CoRR  \textbf{abs/2205.07246} (2022)

\bibitem{DBLP:conf/iclr/0005W0u21}
Wei, C., Wang, H., Shen, W., Yuille, A.L.: {CO2:} consistent contrast for
  unsupervised visual representation learning. In: {ICLR}. OpenReview.net
  (2021)

\bibitem{DBLP:conf/aaai/YangACX22}
Yang, C., An, Z., Cai, L., Xu, Y.: Mutual contrastive learning for visual
  representation learning. In: {AAAI}. pp. 3045--3053. {AAAI} Press (2022)

\bibitem{DBLP:conf/cvpr/YangZYWD22}
Yang, M., Zhu, Y., Yu, J., Wu, A., Deng, C.: Divide and conquer: Compositional
  experts for generalized novel class discovery. In: {CVPR}. pp. 14248--14257.
  {IEEE} (2022)

\bibitem{DBLP:conf/nips/ZhangWHWWOS21}
Zhang, B., Wang, Y., Hou, W., Wu, H., Wang, J., Okumura, M., Shinozaki, T.:
  Flexmatch: Boosting semi-supervised learning with curriculum pseudo labeling.
  In: NeurIPS. pp. 18408--18419 (2021)

\bibitem{DBLP:conf/eccv/ZhangIE16}
Zhang, R., Isola, P., Efros, A.A.: Colorful image colorization. In: {ECCV}
  {(3)}. Lecture Notes in Computer Science, vol.~9907, pp. 649--666. Springer
  (2016)

\bibitem{DBLP:journals/corr/abs-2305-06144}
Zhao, B., Wen, X., Han, K.: Learning semi-supervised gaussian mixture models
  for generalized category discovery. CoRR  \textbf{abs/2305.06144} (2023)

\bibitem{DBLP:conf/iccv/Zheng0Y0Z0021}
Zheng, M., Wang, F., You, S., Qian, C., Zhang, C., Wang, X., Xu, C.: Weakly
  supervised contrastive learning. In: {ICCV}. pp. 10022--10031. {IEEE} (2021)

\bibitem{DBLP:conf/cvpr/ZhongFRL0S21}
Zhong, Z., Fini, E., Roy, S., Luo, Z., Ricci, E., Sebe, N.: Neighborhood
  contrastive learning for novel class discovery. In: {CVPR}. pp. 10867--10875.
  Computer Vision Foundation / {IEEE} (2021)

\end{thebibliography}
\end{document}